# Observing Trends in Automated Multilingual Media Analysis

**Ralf Steinberger, Aldo Podavini, Alexandra Balahur, Guillaume Jacquet,
Hristo Tanev, Jens Linge, Martin Atkinson, Michele Chinosi,
Vanni Zavarella, Yaniv Steiner, Erik van der Goot**

*European Commission – Joint Research Centre (JRC), Ispra (VA), Italy*

*E-mail: Ralf.Steinberger@jrc.ec.europa.eu ( corresponding author )*

Any large organisation, be it public or private, monitors the media for information to keep abreast of developments in their field of interest, and usually also to become aware of positive or negative opinions expressed towards them. At least for the written media, computer programs have become very efficient at helping the human analysts significantly in their monitoring task by gathering media reports, analysing them, detecting trends and – in some cases – even to issue early warnings or to make predictions of likely future developments. We present here trend recognition-related functionality of the *Europe Media Monitor* (EMM) system, which was developed by the European Commission's *Joint Research Centre* (JRC) for public administrations in the European Union (EU) and beyond. EMM performs large-scale media analysis in up to seventy languages and recognises various types of trends, some of them combining information from news articles written in different languages and from social media posts. EMM also lets users explore the huge amount of multilingual media data through interactive maps and graphs, allowing them to examine the data from various view points and according to multiple criteria. A lot of EMM's functionality is accessibly freely over the internet or via apps for hand-held devices.[1]

## 1 Introduction

Automated Content Analysis (ACA) is likely to be more limited than human intelligence for tasks such as evaluating the relevance of information for a certain purpose, or such as drawing high-level conclusions. Computer programs are also error-prone because human language is inherently ambiguous and text often only makes sense when the meaning of words and sentences is combined with the fundamental world knowledge only people have. However, computers have the advantage that they can easily process more data in a day than a person can read in a life time. Computer programs are particularly useful in application areas with a time component, such as monitoring the live printed online media, because they can ingest the latest news articles as soon as they get published and they can detect changes and recognise and visualise trends. Due to the amount of textual information they can process, computer programs can be used to gain a wider view based on more empirical evidence. These features make ACA applications powerful tools to complement human intelligence.

At least for the written media, the manual paper clipping process of the past – cutting out newspaper articles and combining them into a customised in-house news digest – has to a large extent been replaced by automatic systems. Computers can take over repetitive work such as gathering media reports automatically, categorising them according to multiple categories, grouping related documents, recognising references to persons, organisations and locations in them, etc. Using this filtered and pre-processed data, human analysts can then focus on the more demanding tasks of evaluating the data, selecting the most relevant information and drawing conclusions. The work of analysts will be more efficient if the computer programs can extract more types of information and if the information they can recognise is of a higher level (such as event descriptions: *Who did what to whom, where and when*; e.g. Atkinson et al. 2011, Piskorski et al. 2011a, Tanev et al. 2008). By combining different types of document sources, such as the news and social media, there may be the added benefit of covering complementary information and different information aspects, such as factual event information, opinions and lists of useful internet links (e.g. Tanev et al. 2012). **Trend** recognition is deemed particularly useful as it partially summarises events and it may help users detect hidden developments that can only be seen from a bird's perspective, i.e. by viewing very large amounts of data. Trend visualisations may serve as *early warning* tools, e.g. when certain keywords are suddenly found frequently or when any combination of other text features suddenly changes, compared to the usual average background (e.g. Linge et al. 2009 for the medical domain). Summarisation can also be achieved by intelligently aggregating information from many documents (e.g. Balahur et al. 2010a) or by producing a textual summary of large document collections (e.g. Kabadjov et al. 2013). **Trend prediction** would then be the next logical step: based on regular historical observations specifically co-occurring with certain trends, it should be possible to predict certain trends when the same feature combinations occur again. Such an effort was described by O'Brien (2002) for the challenging domain of conflict and political instability. A major challenge for complex subject domains such as societal conflict or war is that the data needed for making a reliable prediction may simply not exist and/or that some specific factors may decide on whether or not a conflict arises, factors that lie outside the realm of statistical analysis (e.g. the sudden sickness or death of a political leader). In any case, features for predictions should probably include data that can only be found outside the document corpus, such as statistical indicators on the economy and on the society.

---

[1] The EMM applications NewsBrief, the Medical Information System MedISys and NewsExplorer can be accessed via the URL http://emm.newsbrief.eu/overview.html. All URLs were last visited in June 2015.

The main disciplines contributing to ACA are called *computational linguistics, natural language processing, language engineering* or *text mining*. In recent years, this field has made a leap forward due to insights and methods developed in statistics and in machine learning, and of course due to the strong increase of computer power, the availability of large collections of machine-readable documents and the existence of the internet.

In Section 2, we will give an overview of EMM, its functionality and its users. We will particularly point out the usefulness of aggregating information derived from the news in many different languages. This has the advantage that it reduces any national bias and that it benefits from information complementarity observed in media sources written in different languages. In Section 3, we will then present a variety of trend presentations and data visualisation techniques used in EMM. These include bar graphs and pie charts to visualise feature distributions, maps, trend graphs for reporting intensity and sentiment, early warning visualisations to alert users in a timely fashion, and social network graphs for the visual display of relations between people.

In Section 4, we summarise the benefits of automatic media monitoring. We point out limitations of ACA and the potential dangers of relying on automatically derived information based on large volumes of textual data. Finally, we highlight the importance of drill-down functionality of any automatic system as this allows users to check and verify the accuracy of any automatic analysis.

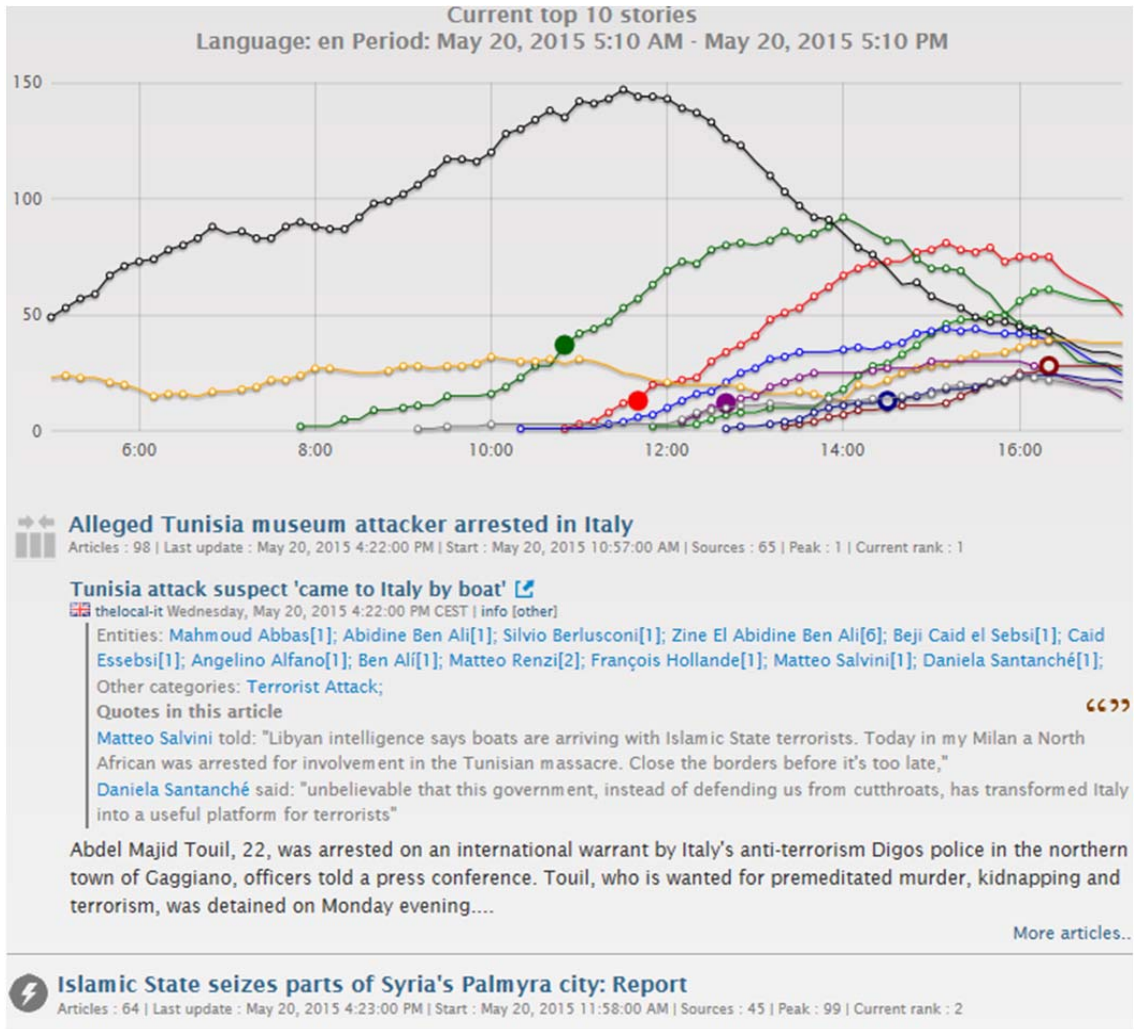

**Figure 1.** Screenshot of EMM-NewsBrief showing the ten most-reported news stories at the moment and their development over time, together with meta-information on the first news cluster (persons mentioned, quotations, first article found, etc.)

## 2  Europe Media Monitor (EMM) – A brief Overview

### 2.1  Overview

*Europe Media Monitor* (EMM) stands for a whole family of media gathering and analysis applications, including *NewsBrief*, *NewsExplorer*, the *Medical Information System* MedISys, *NewsDesk* and more (Steinberger et al. 2009a). EMM was entirely developed at the JRC. While the main **users** are the EU institutions and the national authorities of the 28 EU member states, EMM was also made accessible to international organisations (e.g. various *United Nations* sub-organisations, the *African Union* and the *Organisation of American States*) and to the national authorities of selected partner countries of the EU.

The first version of *NewsBrief* came online in 2002 while NewsExplorer came in 2004, but both systems processed smaller volumes of news and they had less functionality. EMM currently gathers a daily average of about **220,000 online news** articles per day in seventy languages from approximately 5,000 different web sources (status May 2015). The news sources were manually selected with the purpose to represent the major newspapers of all countries in the world and to include European-language news (especially English) from around the world. For reasons of balance, it was decided not to include all easily accessible news sources, but to monitor a comparable number of news sources per country, with a focus on Europe. EMM additionally processes news feeds from over twenty press agencies. It visits news-like websites such as governmental and non-governmental web pages and it monitors social media such as Twitter and FaceBook. The public versions of EMM (see Footnote 1) do not show commercially acquired documents and they offer less functionality than the EC-internal versions.

Separately for each language, the news articles then undergo a series of **processing steps**, including language recognition, document duplicate detection, Named Entity Recognition (NER) for persons, organisations and locations, quotation extraction, sentiment/tonality analysis, and categorisation into one or more of over 1,000 different subject domains. EMM then clusters related articles into groups, which allows users to examine the load of articles in an organised fashion. The different EMM applications provide different functionality, described in the next sub-section.

## 2.2 Family of EMM news monitoring applications

**NewsBrief** (see **Figure 1**) is the most widely used system. It provides users with near-real-time information on their field of interest in all seventy EMM languages. Separately for each language, news gathered within a sliding four-hour window (8 hours for some languages) are grouped into clusters, but older articles remain linked to the cluster as long as new articles arrive. For each cluster, automatically extracted meta-information such as named entities and quotations are displayed. Continuously updated graphs show the ten currently largest clusters and their development over time. By clicking on any of the clusters, users can see the list of all articles and click on each article to read the entire text on the website where it was originally found. For fourteen languages, an automatically pre-generated translation into English is available (Turchi et al., 2012). For event types with relevance to health, safety and security, NewsBrief also displays automatically extracted *event* information (eight languages only), including the event type, location and time of the event, number and type of victims (dead, injured, infected), and – where this was mentioned – the perpetrator (the person or group inflicting the damage) (Tanev et al. 2008; Atkinson et al. 2010, 2011 and 2013). The limitation of the event types is due to the user groups, who are mostly concerned with providing support in case of disasters, epidemics, etc. NewsBrief offers subscriptions for automatic updates per category by email, for institutional users also via SMS.

**MedISys** is rather similar to NewsBrief, except that all its content categories are related to issues that are relevant for Public Health monitoring. Its news categories include all major communicable diseases and other Chemical, Biological, Radiological or Nuclear (CBRN) dangers, symptoms, as well as subjects of scientific or societal value such as vaccinations and genetically modified organisms.

**NewsExplorer** provides a more long-term view of the news (in 21 languages only) and it provides a cross-lingual functionality. Rather than displaying and grouping the current news, NewsExplorer clusters the news of a whole calendar day and displays the clusters ordered by size. For each cluster, hyperlinks lead users to the equivalent news clusters in any of the other twenty languages (where applicable) and to historically related news. NewsExplorer also includes hundreds of thousands of entity pages (persons, organisations and more), where historically gathered information on each entity is aggregated and displayed, including name variants, titles, clusters and quotes where the entity was mentioned, quotes issued by that person, other entities frequently mentioned together with this entity, and more (see **Figure 2**).

**Figure 2**. NewsExplorer entity page, showing the most recent news articles mentioning that person, name spelling variants, titles found in the news, latest news articles and quotations by and about that person. The right-hand-side columns display other entities frequently mentioned together with that person.

**NewsDesk** is a tool for human moderation. It allows media monitoring professionals to view and select the automatically pre-processed news data and to easily create readily formatted in-house newsletters. It is the main tool for the creation of information products by EMM users.

**EMM Apps** for mobile devices such as iOS and Android phones and tablets (see **Figure 3**) became publicly and freely available in 2013. Due to the personal nature of such devices, it became first possible to display customised starting pages for each user. For the iOS EMM App alone, about 26,000 downloads were recorded up to May 2015. This customisable version of EMM became very popular so that this functionality was implemented in a new web version of EMM, called *MyNews* (see below).

The EMM App uses a whole new concept and way to interact with EMM meta-data, referred to as *Channels*. A channel is a stream of EMM articles that all share the same metadata: Channels can be (a) any news category, (b) the top 20 stories in a particular language, (c) a country/category combination, (d) an entity recognised by EMM, or (e) a search in the full-text index. Users can create such channels for themselves and they can group channels into sets, allowing them to browse freely between channels in any of these sets. When users open a channel, they get access to all the articles that are present in the channel at the time, plus the other meta-data that EMM has identified and associated to that channel. Users can of course also browse the attached meta-data, turn them into new channels and *pin* them to the current set.

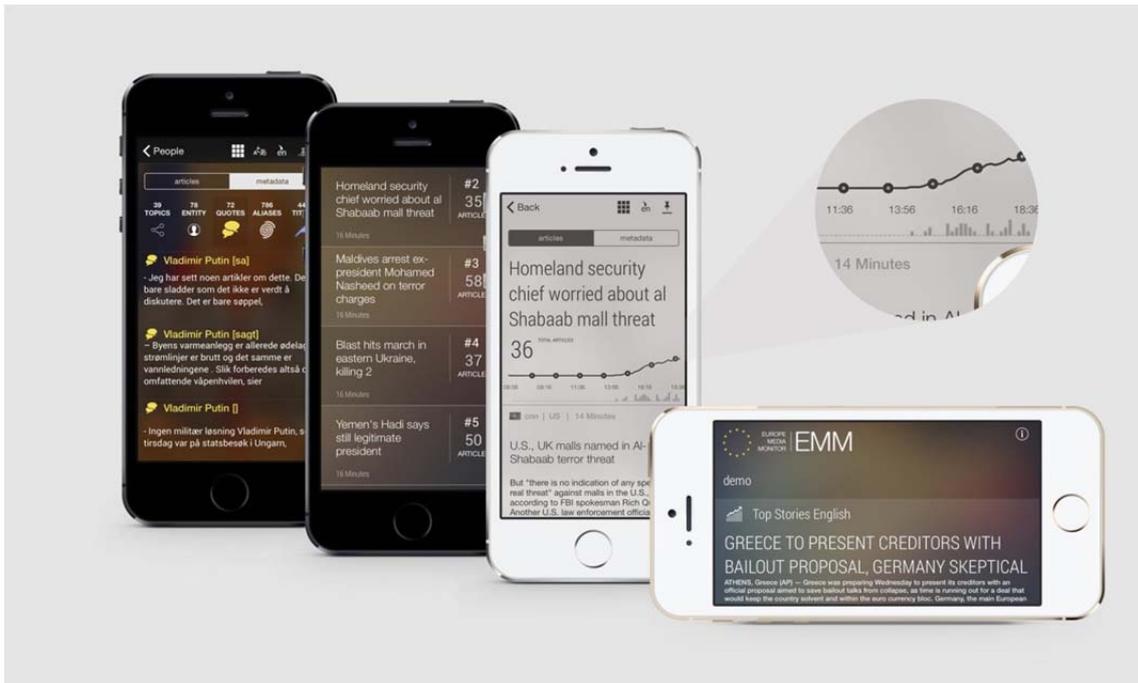

**Figure 3.** Several views of EMM Apps on the iPhone.

Crisis management tools and products have been found to be challenging to design and produce due to the complexity of dynamic customisable data sets defined by each individual user. The main problems in designing such tools are ambiguity, multi-platform support, data representation and other pitfalls commonly seen in mobile technology development. We adhere to a model-based methodology focusing on core functionality and logical interactions with the data set, user-centric design and data visualisation while supporting other development activities including a requirement analysis for a wide set of devices and operating systems, verification and validation. The result of the development cycle is a layout structure in which a wide scale of EMM crisis management tools has been developed.

There are many digital solutions aiming to support humanitarian and emergency response tools by means of open source information gathering and text analysis. A strong trend among them is the ability to detect and analyse vast amounts of data, highlighting important developments relevant to each user and use. Many solutions are already operational today. The majority of these solutions requires the user to open a webpage a few times every day to get updated. Other solutions are relying on communicating with external servers, which is expensive and demanding in maintenance. They additionally usually require user authentication, which can compromise privacy and security.

Our own solution allows custom notifications based on changes in the specific data set the user has defined. When a logical threshold is activated the system displays a notification directly on the user's mobile device. By merging our notifications with the core system notification system of the mobile device, we alert the user only when it is appropriate. For example, notification will wait silently when the user is asleep and will schedule the notifications to be presented a few minutes after the user has started using the device. This is being done without any user intervention or pre-settings. This novel solution differentiates itself from most notification solutions in the fact that it does not rely on any server side technology. The application itself calculates when and how notifications are presented to the user based on an internal logic crossed with background fetching of the current total data set.

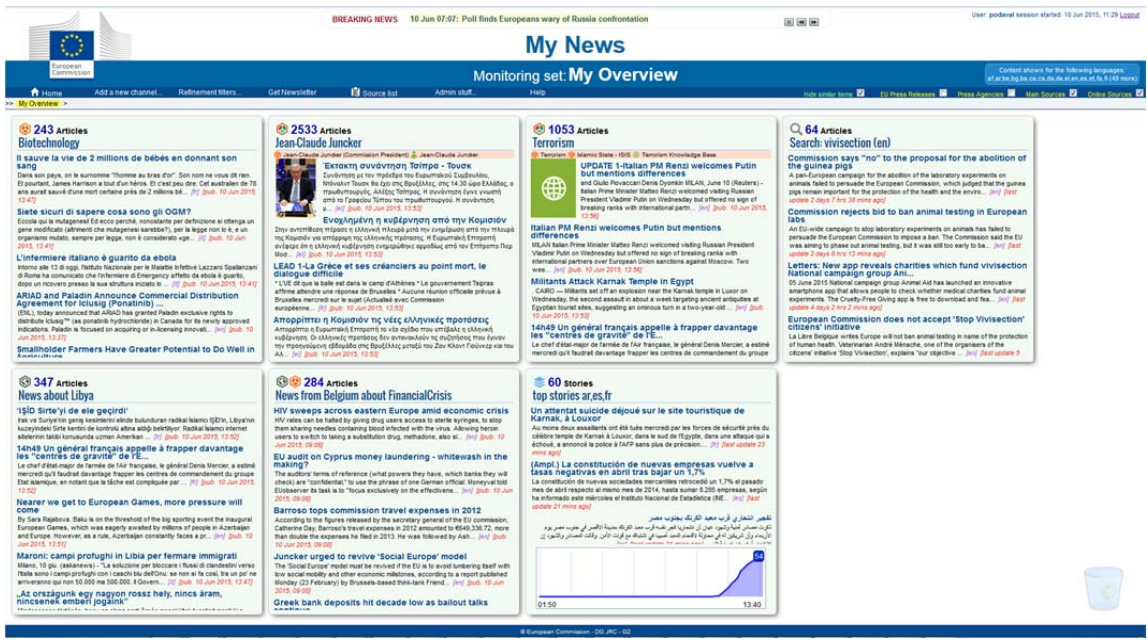

**Figure 4.** Sample page of the customisable *MyNews* application, showing news on subject domains, persons, countries, top news and news from a certain country. Users can thus compile their own newspaper by pinning all themes of their interest to a set.

**MyNews** is the first customisable web interface to the news items supplied by the EMM engine designed for desktop browsers. It became available in 2015. It requires logging in and is only available in-house, i.e. it is not accessible to the wider public. MyNews is highly customisable, since it allows users to define their own specific view by selecting the topics they are most interested in. This is achieved – similarly to the EMM mobile apps – by allowing users to tune news channels focused on very specific topics. They can create as many channels as they like, and they can organise them into sets (see **Figure 4**). There are many different ways to create new channels, which increases greatly the flexibility of the tool, combining as a union or as an intersection of article selections based on (a) text language, (b) news categories, (c) entities, (d) news from a certain country or (e) news about a certain country, (f) top stories (i.e. the biggest clusters of news talking about the same event) or (g) freely chosen search words. When visualising the contents of any of the channels, the meta-data relating specifically to this selection of news is displayed visually (see **Figure 5**).

The **Big Screen App**, available since 2014, offers a view of EMM that is visible on large screens in central locations at user organisations. It shows a revolving and continuously updated view of what is happening around the world, targeted to the respective user communities, using text, maps and graphs.

**Citizens and Science (CAS)** is a project that aims to gauge the relative importance of reporting on Science & Technology (S&T) in traditional and social media (Van der Goot et al. 2015). It does this by comparing the reporting volume from a number of European Nations and the USA of items that correspond to a number of predefined S&T categories. The sources of these items are taken from the traditional online news media, public posts from FaceBook and tweets from Twitter. CAS allows investigating the relative dominance of certain themes across different media (traditional vs. social), languages and countries and it can help find empirical evidence of biased reporting. The data produced in this project raises interesting questions regarding the relative awareness of S&T issues and the dominance of reporting in different countries.

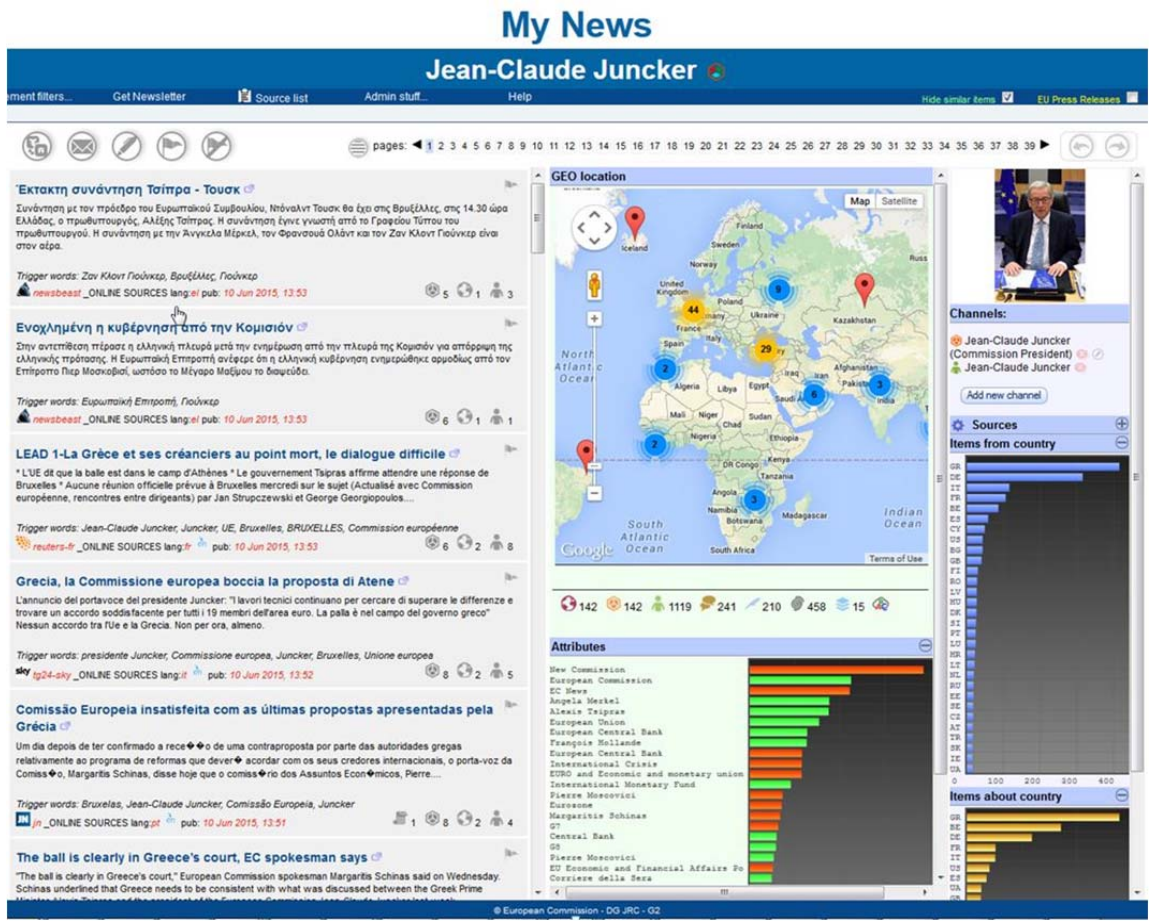

**Figure 5.** View of the channel 'Jean-Claude Juncker', which also includes news articles where the name is spelled differently. The left column shows the list of most recent articles. The middle column shows the locations mentioned in the selected news articles. By zooming in or out on the map, the granularity of the summary level changes. The orange and green bars show the frequency-ranked categories and entities of this set of articles. The graphs on the right visualise the geographical provenance (blue) and the geographical coverage of the news (yellow) displayed on this page.

### 2.3  Multilinguality in EMM

Multilinguality is an extremely important feature in this news monitoring application. Covering so many languages is not only important because the European Union consists of 28 Member States with 24 official EU languages. The coverage of news in 70 different languages is also due to the insight that news reporting is complementary across different countries and languages, both regarding the contents and the opinions expressed in the media. By gathering and analysing different languages, EMM reduces any national or regional bias and it increases the coverage of facts and of opinions.

While major world events such as large-scale disasters, major sports events, wars and meetings of world leaders are usually also reported in English, there is ample evidence that only a minority of the smaller events is reported on in the press outside the country where the event happens. Many EMM users have specialised interests such as the monitoring of events that may have negative effects on Public Health (e.g. disease outbreaks, reports on food poisoning, lack of access to medicines; Linge et al. 2011) or on the stability or welfare of a country (e.g. clashes between ethnic groups, accidents, crime; Atkinson et al. 2010). An analysis has shown that the vast majority of such events is not translated or reported abroad (Piskorski et al. 2011b). The links between related clusters across different languages in NewsExplorer show that only some of the news items in each country or language have an equivalent in other languages while the majority of news clusters talk about subjects of national interest.[2] **Figure 6**, taken from the live EMM news cluster world map[3], also gives evidence of the uneven distribution of language reporting for locations on the globe: News mentioning locations in Latin America are mostly reported in Spanish and Portuguese; there is little news on Russia and China that is not written in Russian or Chinese, respectively, etc. Only by combining the world news in all different languages do we get a fuller picture of what is happening. In order to process news in different languages, the text mining software first has to be developed for these languages. Steinberger (2012) gives an overview of various approaches to minimise the effort to achieve such an extensive language coverage.

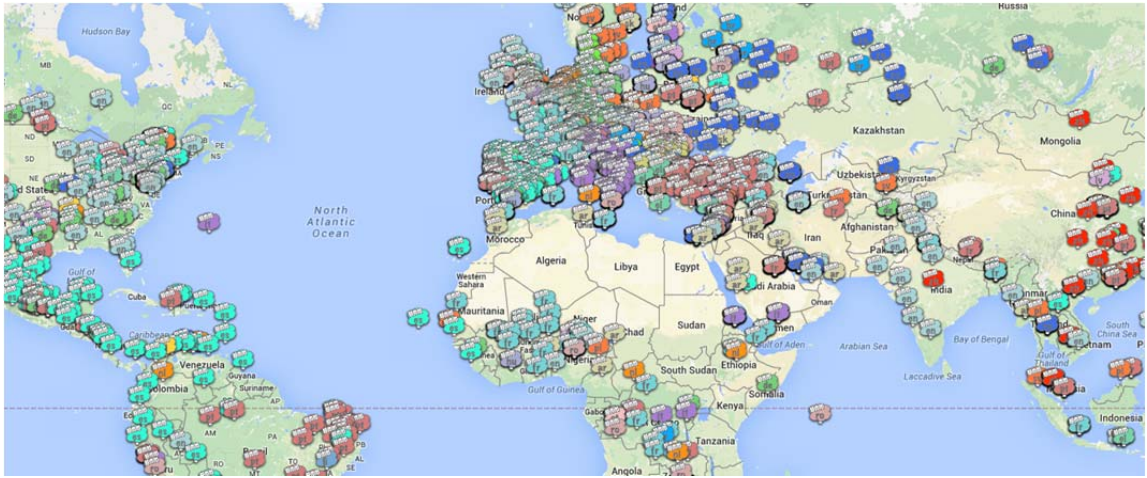

**Figure 6.** NewsBrief cluster map showing which parts of the world are currently most talked about, and in which languages the reports were found. Clouds represent news clusters consisting of at least two news articles. Colours and the corresponding language code represent the publication language of the news articles. The colour distribution shows how some world regions are mostly reported on in one language.

## 3  Trend observation and distribution statistics in EMM

In this section, we want to give some concrete examples of trend monitoring, as well as of bird's views of large amounts of media data giving insights in the relative distribution of news contents. The selection of examples shown here is based on wanting to present different visualisation principles or types, but it is naturally also driven by the interests of EMM users. Since EMM monitors in near-real time (time stamp) large amounts of media reports from around the world and it keeps track of the information (e.g. news provenance, news source, publication language, URL, media type, time of publication, etc.) and it additionally extracts categories and features (e.g. subject domain; number of related articles; names of persons, organisations and locations; sentiment; combinations of features; average values, etc.), it is in principle possible to produce and visualise statistics on *any feature or feature combination*. This can be done for a specific point in time (most EMM users are interested in the *current situation*, i.e. *now*), it can be done for any moment back in time, it is possible to compare current values to average values, and it is possible to perform a time series analysis, i.e. we can show any change over time. Note, however, that – while all such *meta-data* extracted by EMM can be stored – the original full text of the news has to be deleted after the analysis, for copyright reasons. Users will thus be able to see the meta data and a snippet of the news text (title and the first few words), but if they want to see the full text, they have to follow the hyperlink provided. Whether or not the full text is still accessible then depends on the news provider. In the following sub-sections, we will present some types of trend observations and visual presentations of distribution statistics.

---

[2] In EMM-NewsExplorer (http://emm.newsexplorer.eu/NewsExplorer/home/ru/latest.html), the largest news items usually are linked across languages, while many smaller news items are typically of national interest.
[3] See http://emm.newsbrief.eu/geo?format=html&type=cluster&language=all for a live view of this map.

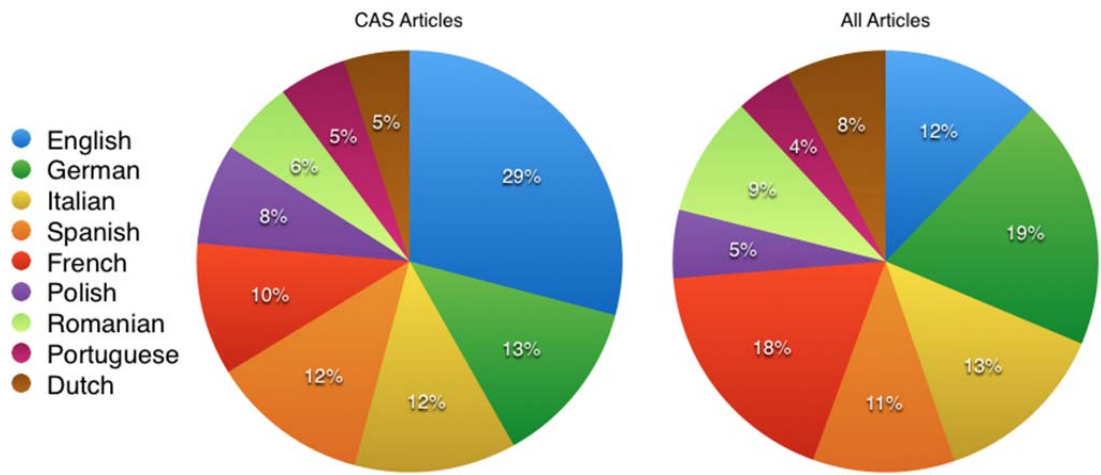

**Figure 7.** Pie charts visualising the language distribution of all European news articles (right) compared to the language distribution for European articles on S&T, for a selection of nine languages.

### 3.1 Bar graphs and pie charts
The simplest and probably clearest way of presenting static data is achieved using bar graphs and pie charts. **Figure 5** shows three different bar charts to visualise different aspects for the same selection of news documents (provenance of the news, countries mentioned in the articles, and subject domains/entities referred to). These charts give the reader an overview of the whole collection of documents and it thus helps them evaluate and categorise the contents before reading them in detail. **Figure 7** shows the language distribution of a multilingual set of European news articles talking on the subject of Science & Technology and comparing it with the language distribution in all articles covering the same time period. It is immediately visible that English and Polish language articles (left) are over-proportionally talking about S&T, while German and French S&T articles are under-represented, compared to the average.

### 3.2 Maps visualising geographical distributions
Map views are rather popular and intuitive. **Figure 5** shows an aggregated map view (number of articles per continent/country/region, depending on the zoom level) while **Figure 6** shows all news clusters (or those in a selection of languages) seen by EMM at the moment. As a lot of map data is available as ready-made layers (terrain, borders, streets, airports, etc.), map visualisation can be rather powerful. **Figure 8** shows the result of EMM's event recognition software, displayed using Google Earth.[4] Any map data in EMM is hyperlinked to web pages containing the original news articles together with the extracted meta-information so that users can verify the contents by reading the underlying news sources. In order to visualise event locations on a map, place names must first be recognised and disambiguated in the text. Pouliquen et al. (2006) describe an approach on how to distinguish whether a string like *Paris* is a person name (e.g. *Paris Hilton*) or a location, and which of the potentially homographic place names world-wide is being referred to (e.g. distinguishing the capital of France from *Paris* in Texas).

---
[4] To see live event maps, go to http://emm.newsbrief.eu/NewsBrief/eventedition/en/latest_en.html and select the KML output format. It is also possible to display events in all eight languages on the same map.

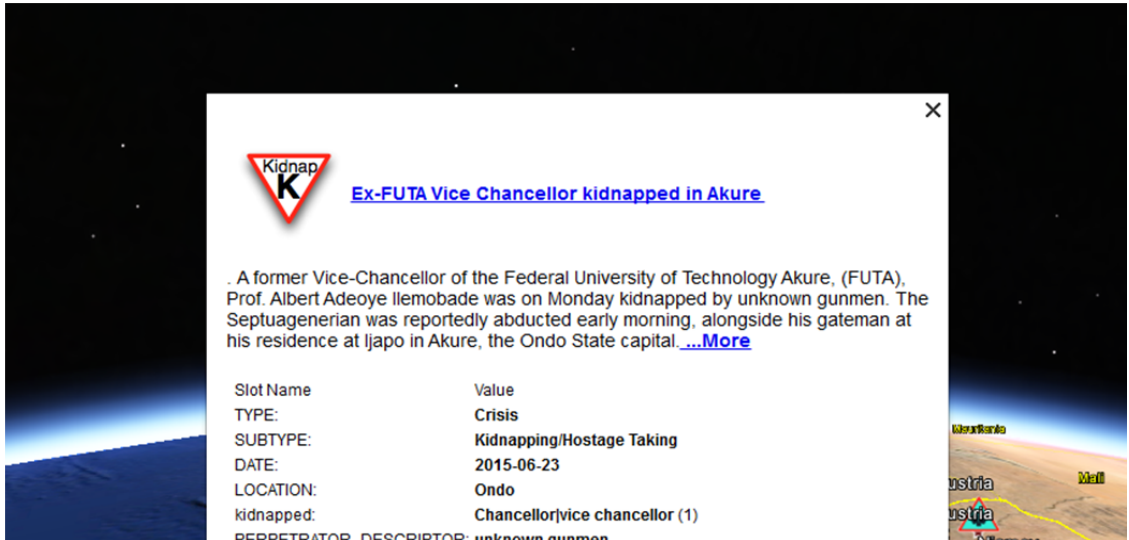

**Figure 9.** Test

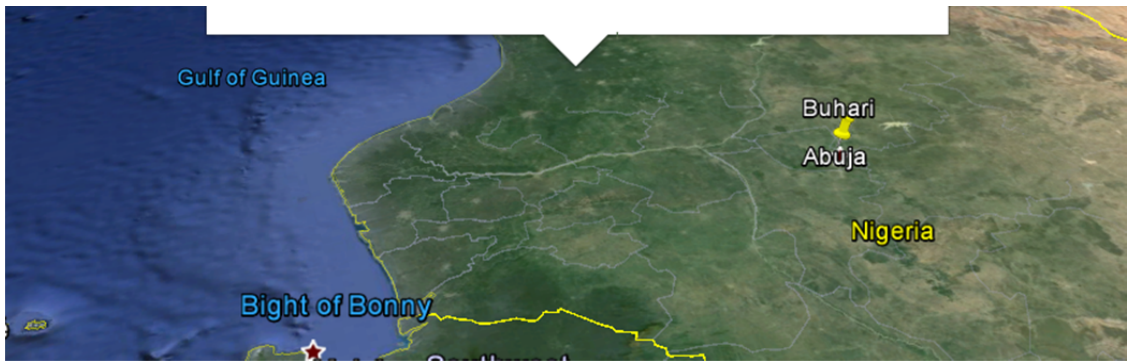

**Figure 8.** Visualisation of automatically extracted event scenario information (who did what to whom, where and when), visualised with Google Maps. This allows showing further information layers such as airports, roads, terrain, borders, etc. By clicking on 'More', users will go to a dedicated page where the original news articles and the related extracted meta-information can be found.

## 3.3 Trend graphs

Trend graphs show a simple correlation between at least two variables, of which one is time. Typically, they take the shape of line graphs or bar graphs where one axis represents time. **Figure 1** shows the size (number of news articles) of the ten largest English language news clusters and their development over the past 12 hours, with a ten-minute resolution (update frequency). The interactive graph clearly shows which stories are most discussed. By hovering with the mouse over any of the points, the most typical news article header of that moment in time is shown so that users can get informed of the development of that story. The system decides on the most typical article header statistically by selecting the medoid, i.e. the document that is closest to the centroid of the vector. By clicking on any of the curves, a new page will open showing the articles that are part of that cluster plus all meta-information available to the system. This graph thus shows ten trend lines in one graph, for the sake of comparison.

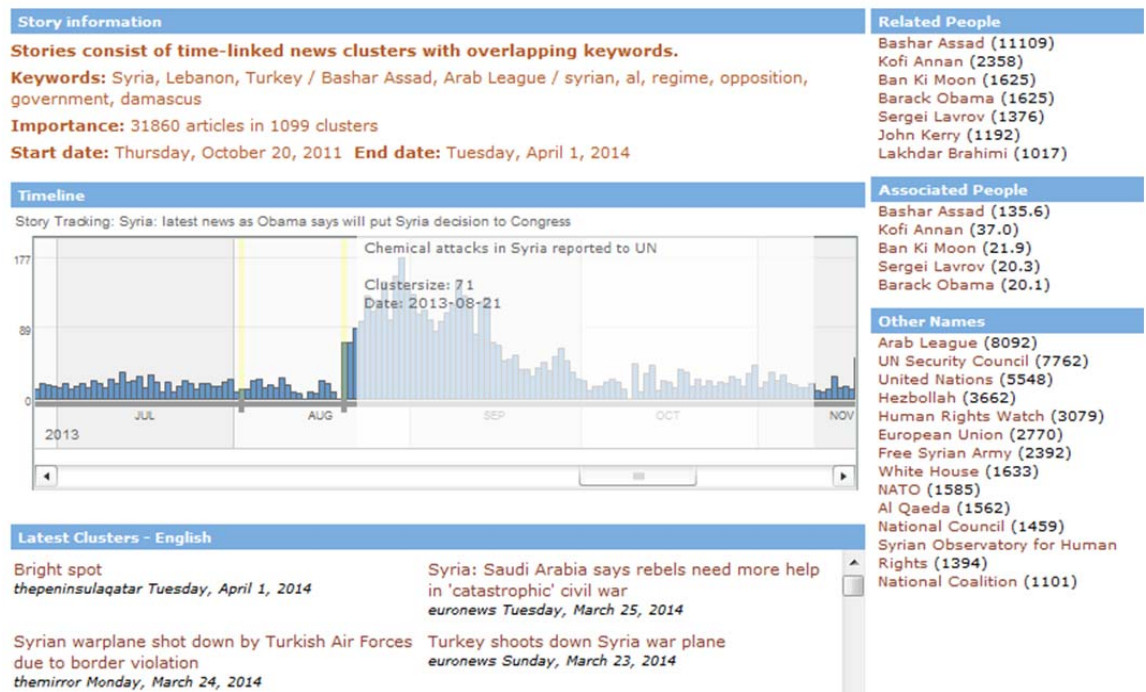

**Figure 10.** Long-term story timeline in EMM-NewsExplorer, showing the intensity of reporting (height of vertical bars) on the Syria conflict and the news cluster title for the day when reporting intensified. The other boxes show meta-information for *all* news articles of the whole reporting period covered in this graph (October 2010-April 2014).

**Figure 10** shows the interactive long-term news story timeline produced in EMM-NewsExplorer. Details are provided in Pouliquen et al. (2008b). The graph shows the number of news articles per day in the daily news clusters about the same event or subject. By hovering over any of the bars, the news cluster title is displayed so that users can explore what happened that day. By clicking on that day, the users are taken to the page with information on that day's news cluster in order to read the articles, see the related meta-information and follow hyperlinks to related reports in other languages. The graph allows exploring developments over longer periods of time and refreshing one's memory on what happened when.

**Figure 11** depicts the development of sentiment automatically detected in Twitter and FaceBook postings (Van der Goot et al. 2015) referring to Information Technology. Sentiment analysis and opinion mining are currently major research areas that lend themselves to a visualisation in trend graphs. Sentiment detection work has been carried out towards entities (Steinberger Josef et al. 2011), in direct speech quotations (Balahur et al. 2009), in social media texts (Balahur & Tanev 2013) and in the traditional printed online media (Balahur et al. 2010).

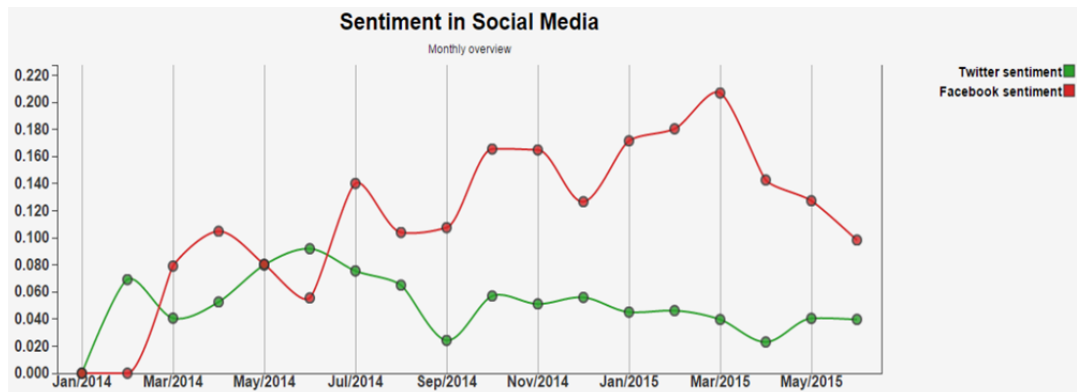

**Figure 11.** Graph showing sentiment detected towards Information Technology-related social media posts over 18 months, using a one-month resolution.

### 3.4 Early warning graphs

**Figure 8** visualises results on the most recent events of a certain type, allowing stakeholders to become aware of the latest developments, to deepen their understanding of what happened (by reading the related news articles) and to take action, if needed.

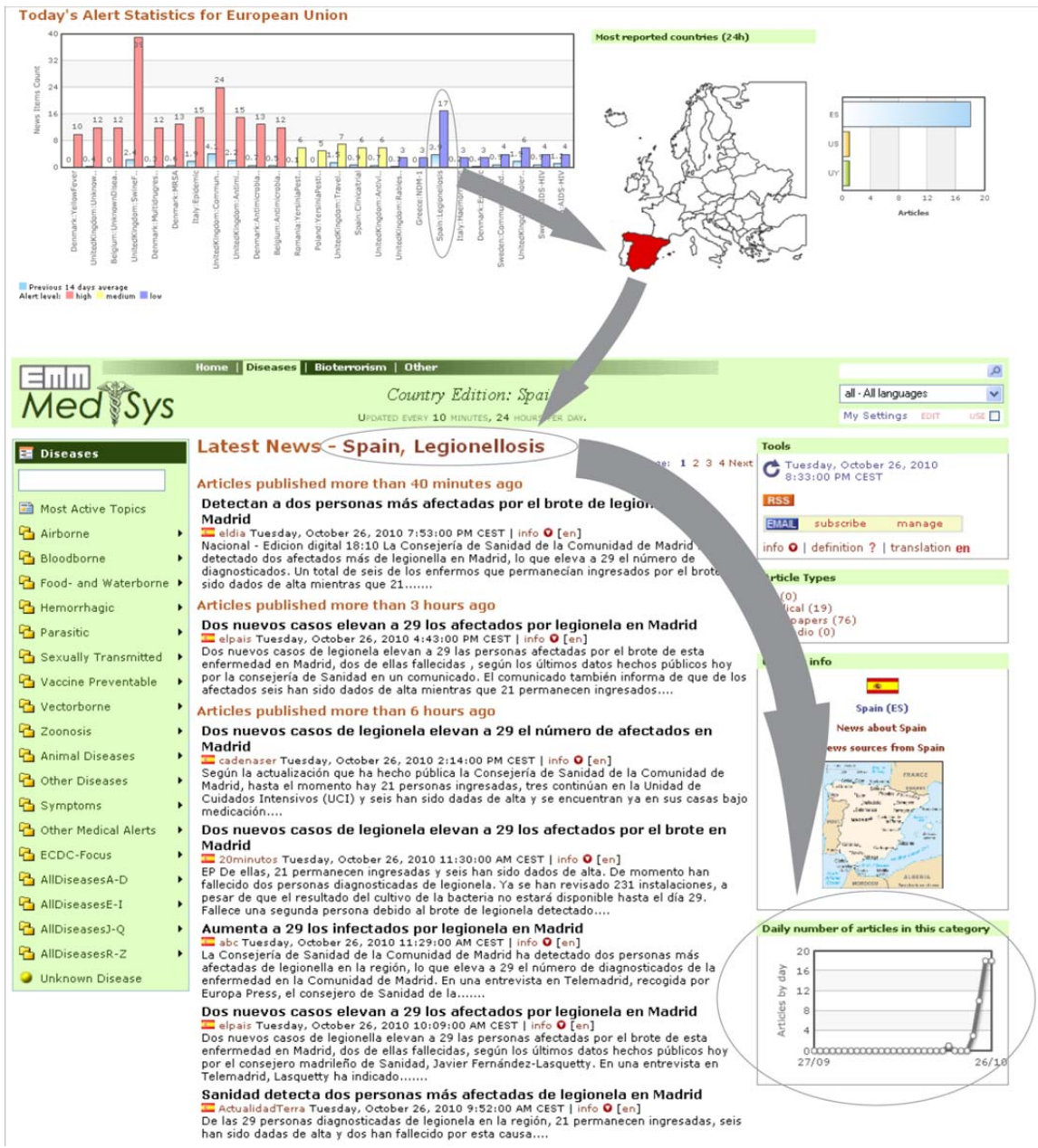

**Figure 12.** Language-independent early warning graph in the *Medical Information System* MedISys (top), displaying the biggest threats to public health at any given moment (country-category combination). By clicking on any bar, users are taken to the web page showing all related news articles plus the meta-information (bottom), which includes a heat map and the two-week development of this category.

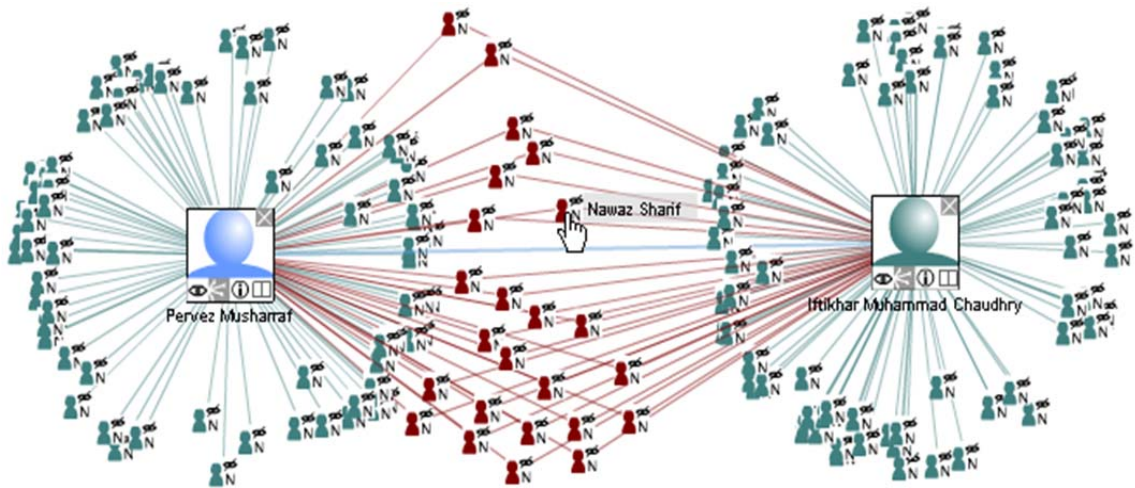

**Figure 13.** The social network node graph taken from EMM-NewsExplorer visualises up to 100 co-occurrence relations per entity for two or more different entities. Relations that are common to at least two entities are highlighted in red. By clicking on any node, users have the option to open the NewsExplorer person page or the Wikipedia page of that entity. Co-occurrences are established even if the names are spelled differently.

Another type of early warning is achieved with statistical means, as shown at the top of **Figure 10**, taken from EMM's *Medical Information System* MedISys (Steinberger et a. 2008), which is used by major public health monitoring organisations around the world (Linge et al. 2011; Barboza et al. 2013). The graph called *daily alert statistics* shows the currently biggest threats world-wide, with decreasing relevance from left to right (the red threats are the ones with the highest alert levels). MedISys counts the number of articles in the last 24 hours for any country-threat combination (e.g. *tuberculosis* and *Poland*) and compares it to the two-week average count for this same combination. This ratio is then normalised by the number of articles for different days of the week (there are less articles on the weekend). The alert statistics graph then shows the results of all calculations, ranked by the value of this ratio. Note that the ratio is entirely independent of the absolute numbers as it rather measures the unexpectedness. Each country-threat combination is shown in two columns: the left one (light blue) shows the observed number of articles while the right one (red, yellow or blue) shows the expected two-week average. An important feature of this graph and of MedISys/EMM as a whole is that this alert is *language-independent*. The same categories for countries and for threats exist for (almost) all EMM languages, meaning that the articles for this threat category may be found in one language only (e.g. Polish or Arabic), which often is different from the languages spoken by the MedISys user. The graph is interactive: Users can click on any of the bars to jump to a new page where all relevant articles for this country-threat combination are displayed, together with a heat map and a trend line showing the development over the past 14 days. The *Spain-legionellosis* threat combination no longer is a top threat as it had already been reported on for four days. MedISys additionally monitors the social media and combines information from these with those of the traditional printed online media (Van der Goot et al. 2013).

## 3.5  Further graph types used in EMM

**Figure 13** shows a node graph visualising co-occurrence relations between people.[5] For each person, the 100 most associated entities (persons or organisations) are displayed. The subset of common entities is highlighted in red. The graph is interactive: by clicking on any of the entity nodes, they jump to a page with the news mentioning that entity and displaying all automatically extracted meta-information (e.g. **Figure 2**), or to the Wikipedia page for that entity. Further entities can be added to the same graph. EMM-NewsExplorer produces the correlation data by counting which entities are mentioned together with which other entities in the same news items. In order to suppress media VIPs such as the US president from the purely frequency-based correlation lists (called *related entities* in NewsExplorer), a weighting formula is used that brings those entities to the top that are mostly mentioned together with this person and not so much with other persons (Pouliquen et al. 2006, 2007 and 2008a). The data, referred to in NewsExplorer as *associated entities*, is produced on the basis of mention co-occurrence in the news in 21 different languages, i.e. it is less biased by the reporting language than data produced by a monolingual media monitoring system. As NewsExplorer recognises spelling variants for the same name, even across scripts and languages, the graphs are produced by aggregating data from all EMM languages (Steinberger & Pouliquen 2009). Alternative approaches to creating social network graphs out of media data are described in Tanev (2007) and Tanev et al. (2010).

---

[5] You can explore this functionality on http://emm.newsexplorer.eu/ by clicking on any entity and then on the graph symbol at the top right. Add new names by searching or by clicking on any of the nodes.

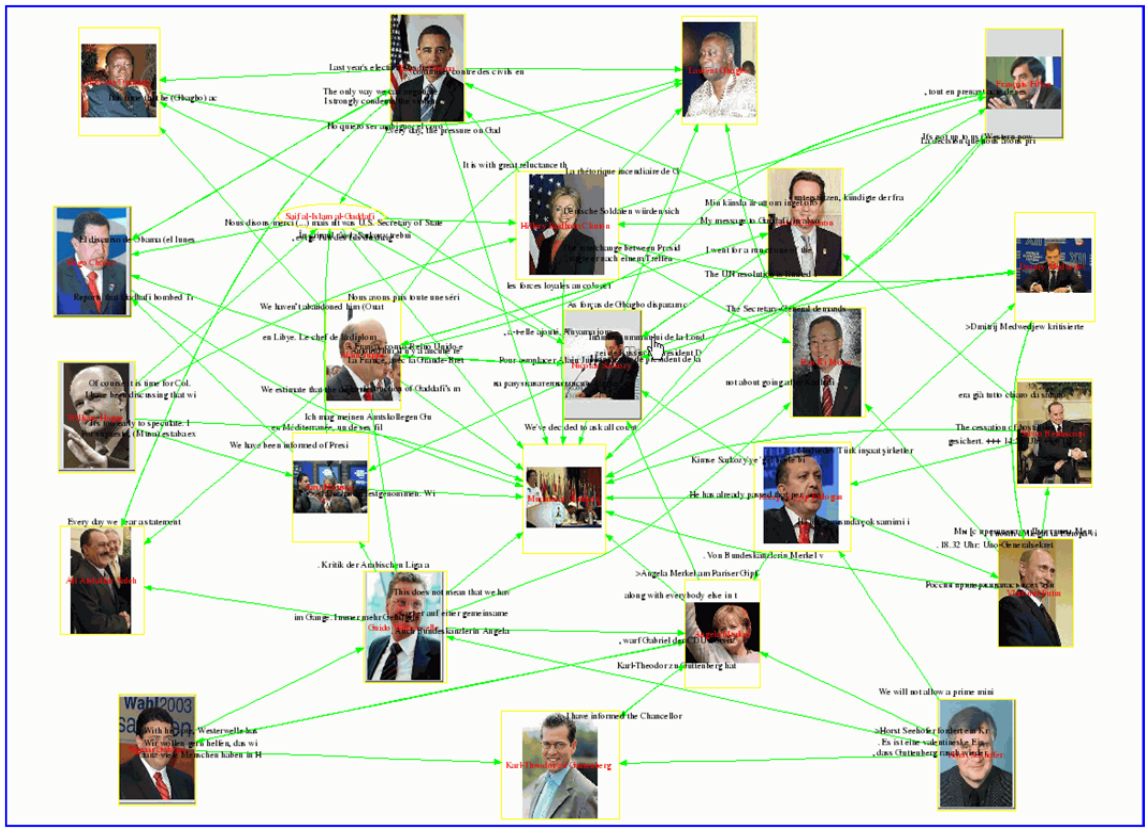

**Figure 14.** Quotation graph visualising who mentions whom in direct speech quotations. People frequently referred to are placed towards the centre of the graph. By clicking on any linking arrow, the automatically extracted quotations are displayed.

EMM recognises direct speech quotations in the news in about twenty different languages and keeps track of who issued the quotation and who is mentioned inside the quotation (Pouliquen et al. 2007). **Figure 14** shows a quotation network indicating who mentions whom (arrows). Persons most referred to are automatically placed closer to the centre of the graph. During the 2007 presidential elections in France, it was observed that Nicolas Sarkozy, who was the winner of the elections, was consistently more central than his opponent Ségolène Royal.

We also observed that – when selecting only news sources from a certain country – the national politicians are usually placed in the centre, meaning that they are frequently quoted and referred to in the national media. Readers who only consume national newspapers will thus usually get an inflated image of the importance of their own politicians on the world-scene.

Quotation networks are no longer used in EMM. The same applies to topic maps, which display the most prominent subject matters referred to in a document collection. The topics are grouped into islands of relatedness (using a method known as Kohonen Maps). The more prominent a group of topics is in the collection, the higher the mountains on the island, with peaks being snow-covered. Just like word clouds, topic maps are pleasant to look at, but it is not so clear how informative they are for the end user.

## 3   Summary and conclusions, pitfalls

Computers have the ability to sieve through large volumes of data in little time and the technologies required for Automated Content Analysis (ACA) have matured to a level where automatically produced results can be useful for the human analyst. We have argued that a man-machine collaboration for the analysis of large volumes of media reports will produce best results because people and computers have complementary strengths. We have presented the main functionality of the European Commission's family of *Europe Media Monitor* (EMM) applications, which currently gathers an average of 220,000 online news articles per day from about 5,000 online news sources in seventy languages (and also from social media postings about certain themes), categorises the news into over 1,000 different categories, groups related articles, extracts various types of information from them, links related articles over time and across languages and presents the analysis results in a variety of ways to the human end user. Moderation tools support the users in viewing the data, in selecting and amending it and in producing in-house newsletters for the information-seeking decision takers. Monitoring not only English or some widely spoken languages is important in order to avoid bias and also because the news is complementary across languages, both for contents and for the sentiment contained therein.

Automatic tools that process and analyse documents turn unstructured information into a structured format that can easily be processed by machines and that also provides useful data for the human user. This results in a data collection where – for each article – we know the news source, the country of origin, the language, the timestamp of the publication, the news categories, the persons, organisations and locations mentioned therein, related articles within the same and across different languages, quotations by and about persons. Additionally, we have data about trends, i.e. whether news related to the same event or subject are increasing or decreasing in numbers over time, and there is some information on sentiment/tonality. This structured collection makes it in principle possible to produce any statistics and to establish any trends related to these types of information. For selected subjects and feature combinations, the JRC regularly publishes its analysis, allowing EMM users to have a deeper insight into the publications on subject areas of their interest. In this article, we presented a range of different types of analyses and visualisations in order to give an overview of distributions and trends observed during large-scale media analysis. Such an extraction and aggregation of data is not usually the final objective, but it normally is the starting point for an intellectual human analysis. Analysts can get inspired by the data, questions may arise, suspicions may get confirmed or contradicted. Used carefully, we believe that the analyses produced by EMM or similar systems can be very useful because they may serve as an inspiration and as empirical evidence for any argument human analysts may want to make. Data produced during ACA can be a good starting point for the more analytic and explanatory work of social and political scientists.

However, we find it extremely important that users be aware of the **limitations** and of **possible pitfalls** when using such data, be it from EMM or from other automatic systems:

(1) First of all, media monitoring is not reality monitoring. What the media say is not necessarily factually true and media attention towards certain subjects usually differs from the real-life distribution of facts or events, giving media consumers a biased view.

(2) Media reporting is heavily influenced by the political or geographical viewpoint of the news source. It is therefore useful to analyse a large, well-balanced set of media sources coming from many different countries world-wide. EMM aims to reach such a balance, but sources are also added on request of users, it is not always known what political standpoints newspapers have, and not all news sources are freely accessible. For this reason, EMM displays the list of media sources monitored so that users can form their own opinion on a possible bias.[6]

(3) Any analysis, be it automatic or man-made, is error-prone. This is even true for basic functionalities such as the recognition of person names in documents and the categorisation of texts according to subject domains. Machines might make simple mistakes easily spottable by human analysts, such as categorising an article as being about the outbreak of communicable diseases when category-defining words such as *tuberculosis* and *sick* are found in articles, for instance talking about a new song produced by a famous music producer. On the other hand, machines are better at going through very large document collections and they are very consistent in their categorisation while people suffer from inconsistency and they tend to generalise on the basis of the small document collection they have read.

(4) For these reasons, it is crucial that any summaries, trend visualisations or other analyses can be verified by the human analysts. Users should be able to view the original data by drilling down, e.g. reading the original text in the case of peaks or unexpected developments, and especially to get an intuitive confidence measure by viewing a number of cases that have led to the conclusions. Most of EMM's graphs are interactive and allow viewing the underlying data. It would be useful if system providers additionally offered confidence values regarding the accuracy of their analyses. For EMM, most specialised publications on individual information extraction tools include such tool evaluation results and an error analysis. However, the tools can behave very differently depending on the text type and the language, making the availability of drill-down functionality indispensable.

(5) End users should be careful with accuracy statistics given by system providers. Especially commercial vendors (but not only) are good at presenting their systems in a very positive light. For instance, our experience has shown that, especially in the field of sentiment analysis (opinion mining, tonality), high accuracy is difficult to achieve even when the statistical accuracy measurements *Precision* and *Recall* are high. Overall Precision (accuracy of the predictions made by the system) may for instance indeed be high when considering predictions for the sum of *positive*, *negative* and *neutral* sentiment, but this might simply be because the majority class (e.g. *neutral*) is very large and the system is good at spotting this. Accuracy statistics may also have been produced on an easy-to-analyse dataset while the data at hand may be harder to analyse.

---

[6] See http://emm.newsbrief.eu/NewsBrief/sourceslist/en/list.html.

> Sentiment may for instance be easier to detect for product review pages on vending sites such as Amazon than on the news because journalists tend to want to give the impression of neutrality.
> (6) Machine learning approaches to text analysis are particularly promising because computers are good at optimising evidence and because machine learning tools are cheap to produce, compared to man-made rules. However, the danger is that the automatically learnt rules are applied to texts that are different from the training data because comparable data rarely exists. Manually produced rules might be easier to tune and to adapt. Again, statistics on the performance of automatic tools should be considered with care. Within EMM, machine learning is used to learn vocabulary and recognition patterns, but these are then usually manually verified and generalised (e.g. Zavarella et al. 2010; Tanev et al. 2009; Tanev & Josef Steinberger 2013).

To summarise: we firmly believe that Automated Content Analysis works when it is used with care and when its advantages and limitations are known. Computers and people have different strengths which – in combination – can be very powerful as they combine large-scale evidence gathering with the intelligence of human judgement.

# 4 References

Atkinson Martin, Jakub Piskorski, Erik van der Goot & Roman Yangarber (2011). Multilingual Real-Time Event Extraction for Border Security Intelligence Gathering. In: U. Kock Wiil (ed.) Counterterrorism and Open Source Intelligence. Springer Lecture Notes in Social Networks, Vol. 2, 1st Edition, 2011, ISBN: 978-3-7091-0387-6, pp 355-390.

Atkinson Martin, Jakub Piskorski, Hristo Tanev, Roman Yangarber & Vanni Zavarella. Techniques for Multilingual Security-related Event Extraction from Online News. In: Przepiórkowski Adam et al. *Computational Linguistics Applications*, pp. 163-186. Springer-Verlag, Berlin, 2013.

Atkinson Martin, Jenya Belayeva, Vanni Zavarella, Jakub Piskorski, S. Huttunen, A. Vihavainen, Roman Yangarber (2010). News Mining for Border Security Intelligence. In IEEE ISI-2010: Intelligence and Security Informatics, Vancouver, BC, Canada.

Balahur Alexandra, Mijail Kabadjov & Josef Steinberger (2010). Exploiting Higher-level Semantic Information for the Opinion-oriented Summarization of Blogs. International Journal of Computational Linguistics and Applications (IJCLA), Vol., No. 1-2, Jan-Dec 2010, pp. 45-59.

Balahur Alexandra & Hristo Tanev (2013). Detecting event-related links and sentiments from social media texts. Proceedings of the Conference of the Association for Computational Linguistics (ACL'2013).

Balahur Alexandra, Ralf Steinberger, Erik van der Goot, Bruno Pouliquen & Mijail Kabadjov (2009). Opinion Mining on Newspaper Quotations. Proceedings of the workshop 'Intelligent Analysis and Processing of Web News Content' (IAPWNC), held at the 2009 IEEE/WIC/ACM International Conferences on Web Intelligence and Intelligent Agent Technology, pp. 523-526. Milano, Italy, 15.09.2009.

Balahur Alexandra, Ralf Steinberger, Mijail Kabadjov, Vanni Zavarella, Erik van der Goot, Matina Halkia, Bruno Pouliquen & Jenya Belyaeva (2010). Sentiment Analysis in the News. In: Proceedings of the 7th International Conference on Language Resources and Evaluation (LREC'2010), pp. 2216-2220. Valletta, Malta, 19-21 May 2010.

Barboza P, Vaillant L, Mawudeku A, Nelson NP, Hartley DM, Madoff LC, Linge JP, Collier N, Brownstein JS, Yangarber R, Astagneau P (2013). Early Alerting Reporting Project Of The Global Health Security Initiative. Evaluation of epidemic intelligence systems integrated in the early alerting and reporting project for the detection of A/H5N1 influenza events. PLoS One. 2013;8(3):e57252. doi: 10.1371/journal.pone.0057252. Epub 2013 Mar 5.

Kabadjov Mijail, Josef Steinberger & Ralf Steinberger (2013). Multilingual Statistical News Summarization. In: Thierry Poibeau, Horacio Saggion, Jakub Piskorski & Roman Yangarber (eds), Multi-source, Multilingual Information Extraction and Summarization, pp. 229-252. Isbn: 978-3-642-28569-1, Doi: 10.1007/978-3-642-28569-1_11, Springer, Berlin & Heidelberg, Germany.

Linge Jens, Ralf Steinberger, Thomas Weber, Roman Yangarber, Erik van der Goot, Delilah Al Khudhairy & Nikolaos Stilianakis (2009). Internet Surveillance Systems for Early Alerting of Health Threats. EuroSurveillance Vol. 14, Issue 13. Stockholm, 2 April 2009.

Linge, J.P., Mantero, J. Fuart, F., Belyaeva, J., Atkinson, M., van der Goot, E. (2011). Tracking Media Reports on the Shiga toxin-producing Escherichia coli O104:H4 outbreak in Germany. In: Malaga. P. Kostkova, M. Szomszor, and D. Fowler (eds.), Proceedings of eHealth conference (eHealth 2011), LNICST 91, pp. 178–185, 2012. PUBSY JRC65929.

O'Brien Sean P. (2002). Anticipating the Good, the Bad, and the Ugly. An Early Warning Approach to Conflict and Instability Analysis. Journal of Conflict Resolution, Vol. 46 No. 6, December 2002, pp. 791-811

Piskorski Jakub, Hristo Tanev, Martin Atkinson, Erik van der Goot & Vanni Zavarella (2011). Online News Event Extraction for Global Crisis Surveillance. Transactions on Computational Collective Intelligence. Springer Lecture Notes in Computer Science LNCS 6910/2011, pp. 182-212.

Piskorski Jakub, Jenya Belyaeva & Martin Atkinson (2011). Exploring the usefulness of cross-lingual information fusion for refining real-time news event extraction. Proceedings of the 8th International Conference Recent Advances in Natural Language Processing (RANLP'2011), pp. 210-217. Hissar, Bulgaria, 12-14 September 2011

Pouliquen Bruno, Hristo Tanev & Martin Atkinson (2008). Extracting and Learning Social Networks out of Multilingual News. Proceedings of the social networks and application tools workshop (SocNet-08) pp. 13-16. Skalica, Slovakia, 19-21 September 2008.

Pouliquen Bruno, Marco Kimler, Ralf Steinberger, Camelia Ignat, Tamara Oellinger, Ken Blackler, Flavio Fuart, Wajdi Zaghouani, Anna Widiger, Ann-Charlotte Forslund, Clive Best (2006). Geocoding multilingual texts: Recognition, Disambiguation and Visualisation. Proceedings of the 5th International Conference on Language Resources and Evaluation (LREC'2006), pp. 53-58. Genoa, Italy, 24-26 May 2006.

Pouliquen Bruno, Ralf Steinberger & Clive Best (2007). Automatic Detection of Quotations in Multilingual News. In: Proceedings of the International Conference Recent Advances in Natural Language Processing (RANLP'2007), pp. 487-492. Borovets, Bulgaria, 27-29.09.2007.

Pouliquen Bruno, Ralf Steinberger & Olivier Deguernel (2008). Story tracking: linking similar news over time and across languages. In Proceedings of the 2nd workshop *Multi-source Multilingual Information Extraction and Summarization* (MMIES'2008) held at CoLing'2008. Manchester, UK, 23 August 2008.

Pouliquen Bruno, Ralf Steinberger, Camelia Ignat & Tamara Oellinger (2006). Building and displaying name relations using automatic unsupervised analysis of newspaper articles. Proceedings of the 8th International Conference on the Statistical Analysis of Textual Data (JADT'2006). Besançon, 19-21 April 2006.

Pouliquen Bruno, Ralf Steinberger, Jenya Belyaeva (2007). Multilingual multi-document continuously updated social networks. Proceedings of the Workshop *Multi-source Multilingual Information Extraction and Summarization* (MMIES'2007) held at RANLP'2007, pp. 25-32. Borovets, Bulgaria, 26 September 2007.

Steinberger Josef, Polina Lenkova, Mijail Kabadjov, Ralf Steinberger & Erik van der Goot (2011). Multilingual Entity-Centered Sentiment Analysis Evaluated by Parallel Corpora. Proceedings of the 8th International Conference Recent Advances in Natural Language Processing (RANLP'2011), pp. 770-775. Hissar, Bulgaria, 12-14 September 2011.

Steinberger Ralf & Bruno Pouliquen (2009). Cross-lingual Named Entity Recognition. In: Satoshi Sekine & Elisabete Ranchhod (eds.): Named Entities - Recognition, Classification and Use, Benjamins Current Topics, Volume 19, pp. 137-164. John Benjamins Publishing Company. ISBN 978-90-272-8922 3.

Steinberger Ralf (2012). A survey of methods to ease the development of highly multilingual Text Mining applications. Language Resources and Evaluation Journal, Springer, Volume 46, Issue 2, pp. 155-176 (DOI 10.1007/s10579-011-9165-9).

Steinberger Ralf, Bruno Pouliquen & Erik van der Goot (2009). An Introduction to the Europe Media Monitor Family of Applications. In: Fredric Gey, Noriko Kando & Jussi Karlgren (eds.): Information Access in a Multilingual World - Proceedings of the SIGIR 2009 Workshop (SIGIR-CLIR'2009), pp. 1-8. Boston, USA. 23 July 2009.

Steinberger Ralf, Flavio Fuart, Erik van der Goot, Clive Best, Peter von Etter & Roman Yangarber (2008). Text Mining from the Web for Medical Intelligence. In: Fogelman-Soulié Françoise, Domenico Perrotta, Jakub Piskorski & Ralf Steinberger (eds.): Mining Massive Data Sets for Security. pp. 295-310. IOS Press, Amsterdam, The Netherlands

Tanev Hristo & Josef Steinberger (2013). Semi-automatic acquisition of lexical resources and grammars for event extraction in Bulgarian and Czech. Proceedings of the 4th Biennial International Workshop on Balto-Slavic Natural Language Processing, held at ACL'2013, pp. 110-118.

Tanev Hristo (2007). Unsupervised Learning of Social Networks from a Multiple-Source News Corpus. Proceedings of the Workshop *Multi-source Multilingual Information Extraction and Summarization* (MMIES'2007) held at RANLP'2007, pp. 33-40. Borovets, Bulgaria, 26 September 2007.

Tanev Hristo, Bruno Pouliquen, Vanni Zavarella & Ralf Steinberger (2010). Automatic Expansion of a Social Network Using Sentiment Analysis. In: Nasrullah Memon, Jennifer Jie Xu, David Hicks & Hsinchun Chen (eds). *Annals of Information Systems, Volume 12. Special Issue on Data Mining for Social Network Data, pp. 9-29.* Springer Science and Business Media (DOI 10.1007/978-1-4419-6287-4_2).

Tanev Hristo, Jakub Piskorski & Martin Atkinson (2008). Real-time News Event Extraction for Global Crisis Monitoring. In V. Sugumaran, M. Spiliopoulou, E. Kapetanios (editors) Proceedings of 13th International Conference on Applications of Natural Language to Information Systems (NLDB 2008 ), Lecture Notes in Computer Science, Cool. 5039, 24-27 June, London, UK.

Tanev Hristo, Maud Ehrmann, Jakub Piskorski & Vanni Zavarella (2012). Enhancing Event Descriptions through Twitter Mining. In: AAAI Publications, Sixth International AAAI Conference on Weblogs and Social Media, pp 587-590. Dublin, June 2012.

Tanev Hristo, Vanni Zavarella, Jens Linge, Mijail Kabadjov, Jakub Piskorski, Martin Atkinson & Ralf Steinberger (2009). Exploiting Machine Learning Techniques to Build an Event Extraction System for Portuguese and Spanish. In: linguaMÁTICA Journal:2, pp. 55-66. Available at: http://linguamatica.com/index.php/linguamatica/article/view/37.

Turchi Marco, Martin Atkinson, Alastair Wilcox, Brett Crawley, Stefano Bucci, Ralf Steinberger & Erik van der Goot (2012). ONTS: "OPTIMA" News Translation System. Proceedings of the 13th Conference of the European Chapter of the Association for Computational Linguistics (EACL), pp. 25–30, Avignon, France, April 23 - 27 2012.

Van der Goot Erik, Hristo Tanev & Jens Linge (2013). Combining twitter and media reports on public health events in MedISys. Proceedings of the 22nd international conference on World Wide Web companion, pp. 703-718. International World Wide Web Conferences Steering Committee, 2013.

Van der Goot Erik, Martin Atkinson, Hristo Tanev, Michele Chinosi & Alexandra Balahur (2015). Monitoring and analysis of reporting on Science and Research in the media in the context of "Citizen and Science". JRC Report JRC96546. European Commission, Ispra, Italy.

Zavarella Vanni, Hristo Tanev, Jens Linge, Jakub Piskorski, Martin Atkinson & Ralf Steinberger (2010). Exploiting Multilingual Grammars and Machine Learning Techniques to Build an Event Extraction System for Portuguese. In: Proceedings of the International Conference on Computational Processing of Portuguese Language (PROPOR'2010), Porto Alegre, Brazil, 27-30 April 2010. Springer Lecture Notes for Artificial Intelligence, Vol. 6001, pp. 21-24. Springer.